\pdfoutput=1
\documentclass[11pt]{article}
\usepackage{amsmath, amssymb, amsthm}
\usepackage{graphicx}
\usepackage{hyperref}
\usepackage{geometry}
\usepackage[utf8]{inputenc}
\usepackage{authblk} 
\usepackage{tabularx} % For tables with adjustable column widths
\usepackage{subcaption} % For subfigures (deprecated but still works)
\usepackage{tikz}
\usepackage{pgfplots}
\usepackage{arydshln}

\usepackage[numbers]{natbib} % Numeric style citations

\title{Diffusion Features for Zero-Shot 6DoF Object Pose Estimation}

\author[1,*]{Bernd Von Gimborn}
\author[2]{Philipp Ausserlechner}
\author[2]{Markus Vincze}
\author[1,*]{Stefan Thalhammer}
\affil[1]{Department of Industrial Engineering, University of Applied Sciences Technikum Wien}
\affil[2]{Automation and Control Institute, TU Wien}

\begin{document}
\maketitle
\def\floatpagepagefraction{1}
\def\textpagefraction{.001}

\let\thefootnote\relax\footnotetext{\hspace{-1.8em}$^*$ Equally contributing first authors.}
\let\thefootnote\relax\footnotetext{\hspace{-1.8em}This work has been submitted to Elsevier's *Image and Vision Computing* for possible publication. This work was supported by the EU-program EC Horizon 2020 for Research and Innovation under grant agreement No. 101120823 project MANiBOT funded by the European Union.}

\begin{abstract}
Zero-shot object pose estimation enables the retrieval of object poses from images without necessitating object-specific training. 
In recent approaches this is facilitated by vision foundation models (VFM), which are pre-trained models that are effectively general-purpose feature extractors. 
The characteristics exhibited by these VFMs vary depending on the training data, network architecture, and training paradigm.
The prevailing choice in this field are self-supervised Vision Transformers (ViT).
This study assesses the influence of Latent Diffusion Model (LDM) backbones on zero-shot pose estimation.
In order to facilitate a comparison between the two families of models on a common ground we adopt and modify a recent approach.
Therefore, a template-based multi-staged method for estimating poses in a zero-shot fashion using LDMs is presented.
The efficacy of the proposed approach is empirically evaluated on three standard datasets for object-specific $6DoF$ pose estimation.
The experiments demonstrate an Average Recall improvement of up to $27\%$ over the ViT baseline.
The source code is available at: \href{https://github.com/BvG1993/DZOP}{https://github.com/BvG1993/DZOP}.
\end{abstract}

% need to put that back in again later
%\begin{graphicalabstract}
%\includegraphics{figs/pipeline.png}
%\end{graphicalabstract}

\maketitle

%%%%%%%%%%%%%%%%%%%%%%%%
%%%%%%%%%%%%%%%%%%%%%
% Introduction
%%%%%%%%%%%%%%%%%%%%
%%%%%%%%%%%%%%%%%%%%
\section{Introduction}
Object pose estimation is a fundamental challenge in the fields of computer vision and robotics, as it enables a range of downstream tasks, including object grasping and scene understanding~\cite{liu2023robotic,thalhammer2024challenges}.
Deep learning-based object pose estimation has demonstrated improved pose estimation accuracy, occlusion handling, and runtime as compared to classical approaches that use hand-crafted feature extractors. 
These advances are accompanied by the potential for CNNs to facilitate precise pose estimation from low-cost RGB sensors, while traditional approaches require comparably expensive depth sensors.
The primary disadvantage of RGB-based deep object estimation is the necessity of extensive and diverse training data, which are employed for offline training of instance- or category-specific pose estimators. 
The rendering of $50k$ physically plausible~\cite{pharr2016physically} training images per dataset has been established as the standard case for instance-specific pose estimation~\cite{hodan2024bop}.
This results in training times of up to days per object instance, on modern high-end consumer graphical processing units~\cite{lin2024hipose,thalhammer2024challenges}.
To circumvent these shortcomings, recent research has shifted to zero-shot approaches~\cite{ausserlechner2023zs6d,labbé2022megapose6dposeestimation,ornek2023foundpose,shugurov2022osopmultistageshotobject,wen2024foundationpose}.
In this context, zero-shot refers to the ability to perform tasks on previously unseen data without the need for specific prior training.
This allows the estimation of poses for objects that have not been included in the training process.
These works either employ a single, or multiple real reference images, or images derived from object priors~\cite{ausserlechner2023zs6d,di2024zero123}.
The latter approach yields more accurate results, reflecting the greater coverage of object viewpoints.
The general concept aligns with classical pipelines: a) reference images are generated from mesh priors, b) image-to-image correspondences are estimated, and c) a coarse object pose is derived, with the option of further refinement.

The advent of Vision Foundation Models (VFM) and their capability to generalize to unseen visual domains render them a more powerful alternative to hand-crafted classical descriptors.
The current state of the art for instance-level zero-shot object pose estimation employs Vision Transformers (ViTs) for feature extraction and correspondence matching~\cite{ausserlechner2023zs6d,ornek2023foundpose,wen2024foundationpose}.
A review of the literature on other zero-shot computer vision tasks reveals that these can be performed with either ViTs or Latent Diffusion Models (LDMs). 
This study examines the potential of LDMs as a viable alternative to ViTs for zero-shot object pose estimation, identifying the advantages and disadvantages of this VFM.
In order to achieve this, we present a method for estimating $6DoF$ object poses from an RGB image and a set of object priors.
To directly compare LDMs to ViTs, our empirical analysis is based on ZS6D~\cite{ausserlechner2023zs6d}.
ZS6D employs Dino~\cite{caron2023emerging} for the extraction of features for the matching of object templates to a query image. 
The features of the matched template are then repurposed for the estimation of local correspondences and the derivation of geometric correspondences between the query image and the object prior.
Ultimately, poses are estimated using the Perspective-\textit{n}-Points algorithm~\cite{lepetit2009epnp}.
In this study, we replace the ViT feature extractor in ZS6D with Stable Diffusion (SD)~\cite{Rombach_2022_CVPR}.
To ensure optimal performance with diffusion features, we also modify the downstream pipeline stages, specifically template matching and correspondence estimation.
To elaborate, while the unprocessed SD features are suited for substituting ViT features for template matching, correspondence estimation necessitates a more intricate matching strategy because attention maps are not directly accessible, in contrast to ViTs.
The estimation of robust correspondences requires the use of hyperfeatures~\cite{zhang2023tale}.
The co-projection of the query and template features results in an aligned feature space, which allows for robust corresponding clustering.
This significantly improves the pose estimation accuracy and the runtime.
Furthermore, sub-pixel accurate correspondence estimation stabilizes the pose estimation of objects with a low object-to-image ratio.
A direct comparison is provided on three standard datasets, LMO~\cite{brachmann2014learning}, YCBV~\cite{xiang2018posecnn}, and TLESS~\cite{hodan2017tless}. 
These cover prevalent challenges in object pose estimation, including occlusion, object symmetries, strong illumination changes, and the absence of texture.
The results demonstrate that employing LDMs, as compared to ViT, enhances object pose estimation accuracy across all three datasets, with a maximum of up to $27\%$ relative improvement.

In summary our contributions are:
\begin{enumerate}
    \item A Latent Diffusion feature-based pipeline for zero-shot $6DoF$ object pose estimation.
    \item An empirical evaluation that demonstrates that the features of Latent Diffusion Model are more effective than those of Vision Transformer for the purpose of template matching and for zero-shot $6DoF$ object pose estimation.
    \item A strategy for processing pre-trained diffusion features with the objective of improving the accuracy of correspondence matching.
\end{enumerate}

The remainder of the paper is structured as follows: the Background section provides an overview of recent works on object pose estimation and VFMs; the Method section presents the approach in detail; the Experiments section quantifies and evaluates the findings; and the Discussion section offers a critical reflection on the paper and provides an outlook for future work.

%%%%%%%%%%%%%%%%%%%%
%%%%%%%%%%%%%%%%%%%%
\section{Background}

This section discusses object pose estimation, exploring the recent resurgence of zero-shot object pose estimation enabled by the advent of foundation models.
Additionally, it introduces the concept of VFMs, with a particular focus on the ViT model~\cite{dosovitskiy2021image} and the LDM~\cite{Rombach_2022_CVPR}.
These models are discussed as potential approaches for retrieving generally applicable feature embeddings.

\subsection{Single-shot Object Pose Estimation}

Single-shot object pose estimation refers to the process of inferring the object pose from a single RGB or RGB-D image.
Classical object pose estimation approaches initially compute a pose hypothesis using hand-crafted features to encode a reference view code book~\cite{aldoma2011cad,buch2013pose,drost2010model,hinterstoisser2012model}. During runtime the extracted features from the query image are compared to the code book in order to retrieve pose hypotheses.
In contrast to this strategy, deep object pose estimation generates hypotheses from latent object representations. This, often instance-specific, strategy has been the dominant approach over the past decade due to its accuracy and runtime efficiency~\cite{aing2023faster,jin2024instance,kehl2017ssd,labbe2020cosypose,lin2024hipose,sun2022dynamic,tekin2018real,wang2021gdr,wang2024oa,wang2022multiple,wu2023geometric}.
Instance-specific training necessitates the use of a considerable number of training images, often in the range of thousands, and a training time that can span hours to days~\cite{thalhammer2024challenges,wang2021gdr}.
To overcome these limitation it is necessary to improving the generalization capability of the models and methods in question.
One proposed solution for partially alleviating these issues is the use of category-level approaches, which still necessitate offline training, but demonstrate the ability to generalize to novel objects within known categories~\cite{di2022gpv,remus2023i2c,wang2019normalized}.
Recent studies demonstrate that integrating template-based approaches with deep feature extractors effectively combines the generalizable applicability of classical approaches with the precise pose estimation accuracy and robustness to occlusion observed in deep learning approaches~\cite{ausserlechner2023zs6d,ornek2023foundpose,shugurov2022osopmultistageshotobject,wen2024foundationpose}.
In the context presented here, this is referred to as zero-shot object pose estimation.

\subsection{Zero-shot Pose Estimation}

Classical pose estimation methods use hand-crafted features and descriptors for template matching and correspondence estimation from RGB images.
Recent research has investigated and demonstrated that deep CNN features enhance template matching for pose estimation, surpassing classical approaches~\cite{nguyen2022templates}.
Furthermore, it has been demonstrate that ViTs further enhance template matching\cite{thalhammer2023self}, and are well suited for handling variations between the template and the query image~\cite{goodwin2022}. 
MegaPose~\cite{labbé2022megapose6dposeestimation} and OSOP~\cite{shugurov2022osopmultistageshotobject} demonstrated that conditioning architectures on pose estimation datasets enables novel object $6DoF$ pose estimation. 
In recent works, CNNs have been replaced with self-supervised pre-trained ViTs for zero-shot novel object pose estimation~\cite{ausserlechner2023zs6d,di2024zero123,nguyen2024gigapose,Lin_2024_CVPR,ornek2023foundpose,wen2024foundationpose}.
Additionally, some works investigate LDMs for object pose estimation, focusing on instance-level~\cite{Xu_2024_CVPR} or category-level~\cite{chen2024secondpose} approaches.
This underscores the necessity to assess their viability for zero-shot pose estimation.

\subsection{Vision Foundation Models}

Foundation models are an increasingly prominent area of interest in the fields of computer vision and robotics.
This is due to several factors, including their ability to generalize across tasks, enhance transferability from a limited number of labeled samples, and facilitate zero-shot learning~\cite{caron2023emerging,dosovitskiy2021image,kirillov2023segment,oquab2023dinov2,Rombach_2022_CVPR,zhan2023probing}.
The predominant machine learning architectures that are utilized as VFMs are ViTs and LDMs.
Dosovitskiy~\cite{dosovitskiy2021image} demonstrated that the transformer architecture~\cite{Vaswani2017} originally developed for natural language processing, is also applicable to vision tasks. 
The success of ViTs can be attributed to their diverse advantages over CNNs.
They offer excellent general feature extractors with different training paradigms, such as self-supervised pre-training~\cite{caron2023emerging,oquab2023dinov2}, they provide a unified architecture for vision and language~\cite{zhang2022glipv2}, and they exhibit versatility across different tasks~\cite{kirillov2023segment,vuong2023open,zhang2022glipv2}.
LDMs are based on the denoising probabilistic diffusion process~\cite{ho2020denoising}, which maps the input to a lower-dimensional data representation.
As generative models they synthesize and modify images by iteratively noising and again denoising them. 
They demonstrate particular proficiency in the domains of image synthesis~\cite{Rombach_2022_CVPR}, inpainting~\cite{lugmayr2022repaint}, super-resolution~\cite{gao2023implicit}, and style transfer~\cite{zhang2023inversion}. 
It has been demonstrated that LDMs are excellent zero-shot correspondence matchers~\cite{luo2023dhf,tsagkas2024click}, in some cases similarly effective as ViTs~\cite{zhang2023tale}.
To the best of our knowledge, this work is the first to investigate the suitability of LDMs for zero-shot object pose estimation.

%%%%%%%%%%%%%%%%%%%%%%%%
%%%%%%%%%%%%%%%%%%%%%
% Method
%%%%%%%%%%%%%%%%%%%%
%%%%%%%%%%%%%%%%%%%%
\section{Method}

\begin{figure}
	\centering
	\includegraphics[width=.99\columnwidth]{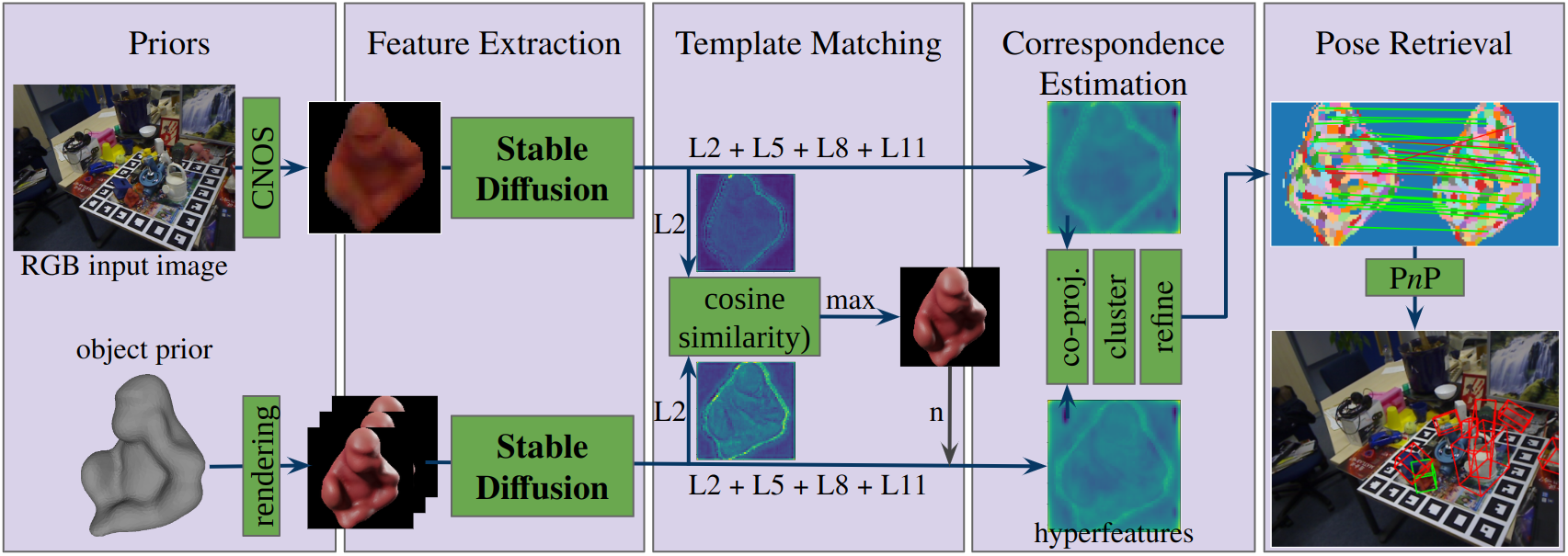}
	\caption{\textbf{DZOP overview} For estimating poses a scene-level RGB image and object meshes of the object of interest are expected. The features of the query and template images are extracted using Stable Diffusion~\cite{Rombach_2022_CVPR}. Templates are matched using the cosine similarity on the feature maps of the second decoder layer of U-net's~\cite{ronneberger2015u} second decoder layer. Semantic correspondences are estimated from clustered hyperfeatures. Ultimately, geometric correspondences are derived and poses are estimated using Perspective-\textit{n}-Points~\cite{hartley2003multiple}.}
	\label{fig:ov}
\end{figure}

This section presents the method for Diffusion-based Zero-shot Object Pose estimation (DZOP) using SD features.
Figure~\ref{fig:ov} illustrates the approach.
As input an RGB image, sparse object location priors in the form of object masks, and priors regarding the objects of interest are required.
To generate object location priors we employ CNOS~\cite{nguyen2023cnosstrongbaselinecadbased}, a zero-shot object segmentation approach that produces objects masks and offers object class hypotheses.
The object priors are employed for generating equidistant object templates, which are rendered from a viewing sphere. 
For feature extraction SD processes the query image and all rendered template images, producing embeddings. 
Template matching is performed on the low-resolution feature maps of the second layer of the U-net backbone's decoder in SD, to determine the optimal template, which is identified as the one exhibiting the highest mutual cosine similarity to the query image.
The process of correspondence estimation employs hyperfeatures, which have been adopted and modified from~\cite{zhang2023tale}, to establish correspondences between the query and the matched template image. 
These hyperfeatures are derived from different feature maps of the U-net's decoder.
Ultimately, the pose is retrieved by deriving geometric correspondences~\cite{park2019pix2pose} from the semantic correspondence locations, and by solving for the $6DoF$ pose using the Perspective-$n$-Points (P\textit{n}P) algorithm~\cite{hartley2003multiple}.

\subsection{Feature Extraction}
Stable Diffusion (SD) employs the iterative denoising process for image generation and manipulation~\cite{ho2020denoising}.
This involves the incremental introduction and removal of noise, resulting in the generation of robust and precise image descriptors. 
The following introduces the fundamental concepts required for an understanding of the diffusion process as it applies for the given task.

\noindent\textbf{Forward diffusion}
The forward diffusion process describes a Markov chain over the time steps $t$. In this process, noise is sequentially added to an image $I$, generating a sequence of increasingly noisy images \( I_1, I_2, \ldots, I_T \), following the formulation below:

\begin{align}
f(I_t \mid I_{t-1}) = \mathcal{N}(I_t; \sqrt{1 - \beta_t} I_{t-1}, \beta_t I)\label{gl:forwarddiffusion}
\end{align}
\noindent where \( \beta_t \) is a time-dependent noise scaling parameter, and \( \mathcal{N} \) represents a normal distribution \cite{ho2020denoising}.

\label{fig:AR_thres}\textbf{Backward diffusion}
The objective of the backward diffusion is to recover the image $I$ from the noisy image $I_t$.
This is achieved by employing a conditioned model \( f_\theta(x_{t-1} \mid x_t) \) that approximates the recursion of the forward diffusion:
\begin{align}
f_\theta(I_{t-1} \mid I_t) = \mathcal{N}(I_{t-1}; \mu_\theta(I_t, t), \Sigma_\theta(I_t, t))\label{gl:backwarddiffusion}
\end{align}
\noindent where \( \mu_\theta \) und \( \Sigma_\theta \) are mean and variance functions, typically encoded with a latent variable model. The backward diffusion is an iterative process that reconstructs $I$ from \( I_T \)~\cite{ho2020denoising}.

\label{fig:AR_thres}\textbf{Stable Diffusion}
Both, the forward and backward diffusion in Equation \ref{gl:forwarddiffusion} and \ref{gl:backwarddiffusion} assume a parametric model for fitting the noise distribution.
We use SD~\cite{Rombach_2022_CVPR}, with a U-Net backbone~\cite{ronneberger2015u}.
The model is pre-trained on a subset of LAION-5B for text-to-image generation~\cite{schuhmann2022laion} using the variational bound loss 
\begin{align}
L_{vb} = \mathbb{E}_q \left[ \sum_{t=1}^T D_{KL}(q(I_{t-1} \mid I_t, I_0) \parallel p_\theta(I_{t-1} \mid I_t)) \right]\label{gl:variational_bound_loss}
\end{align}
\noindent where \( D_{KL} \), the Kullback-Leibler divergence, aligns the ground truth \( f \) and the estimated image distribution \( F_\theta \)~\cite{ho2020denoising}.

\textbf{Diffusion hyperfeatures}
It has been demonstrated that the aggregation of hyperfeatures derived from multiple timesteps and feature maps of the diffusion process leads to enhanced correspondence matching accuracy~\cite{luo2023dhf,zhang2023tale}.
Both parameters influence the captured semantics and thus are relevant to the downstream task.
The authors of ~\cite{luo2023dhf} suggest that utilizing feature maps of the early timesteps of the forward diffusion process facilitates the acquisition of information-rich features.
Moreover, the selection of an appropriate number of timesteps allows for a trade-off between short runtime and image reconstruction quality~\cite{vahdat2021score}.
It has been demonstrated that utilizing $50$ is an appropriate choice for general features~\cite{zhan2023probing}.
With regard to the selection of aggregated feature maps, our findings indicate that for template matching the early layers of the U-net decoder, capturing semantic and overall structure, result in the lowest viewpoint deviation between the query and the template.
However, for correspondence estimation, it is advisable to perform feature aggregation over multiple feature maps that are semantically diverse.

\subsection{Template Matching}

Template matching is employed to ascertain an approximate viewpoint of the object in the template $I_p$ and the query image $I_q$. 
In practice, the initial stage, before obtaining $I_q$, is extracting the sparse location from the input image using a segmentation approach~\cite{ausserlechner2023zs6d}.
Consequently, an estimated object mask is available for use in removing the background of $I_q$.
In order to estimate the approximate viewpoint, a set of $n$ two-dimensional template images $\{ \text{I}^{n}_{p} \mid i = 1, 2, \dots, n \}$ is created from the object prior via rendering~\cite{pharr2016physically}.
When comparing these to $I_q$, the matched template is retrieved by maximizing the mutual cosine similarities~\cite{nguyen2022templates} over the set of template images:

\begin{equation}
I_m = \underset{n}{\arg\max} \frac{e^L(I_q) \cdot e^L(I^{n}_p)}{\|e^L(I_q)\| \|e^L(I^{n}_p)\|} \quad \text{where} \quad L = 2
\end{equation} \label{gl:template_matching}

\noindent where $e^L(.)$ is the obtain feature embedding function from the forward diffusion.
The $L$ indicates the index of the used decoder layer output.
For this stage we use the feature maps extracted by the second decoder layer of the U-Net.

\subsection{Semantic Correspondence Estimation}

\begin{figure}
	\centering
	\includegraphics[width=.9\columnwidth]{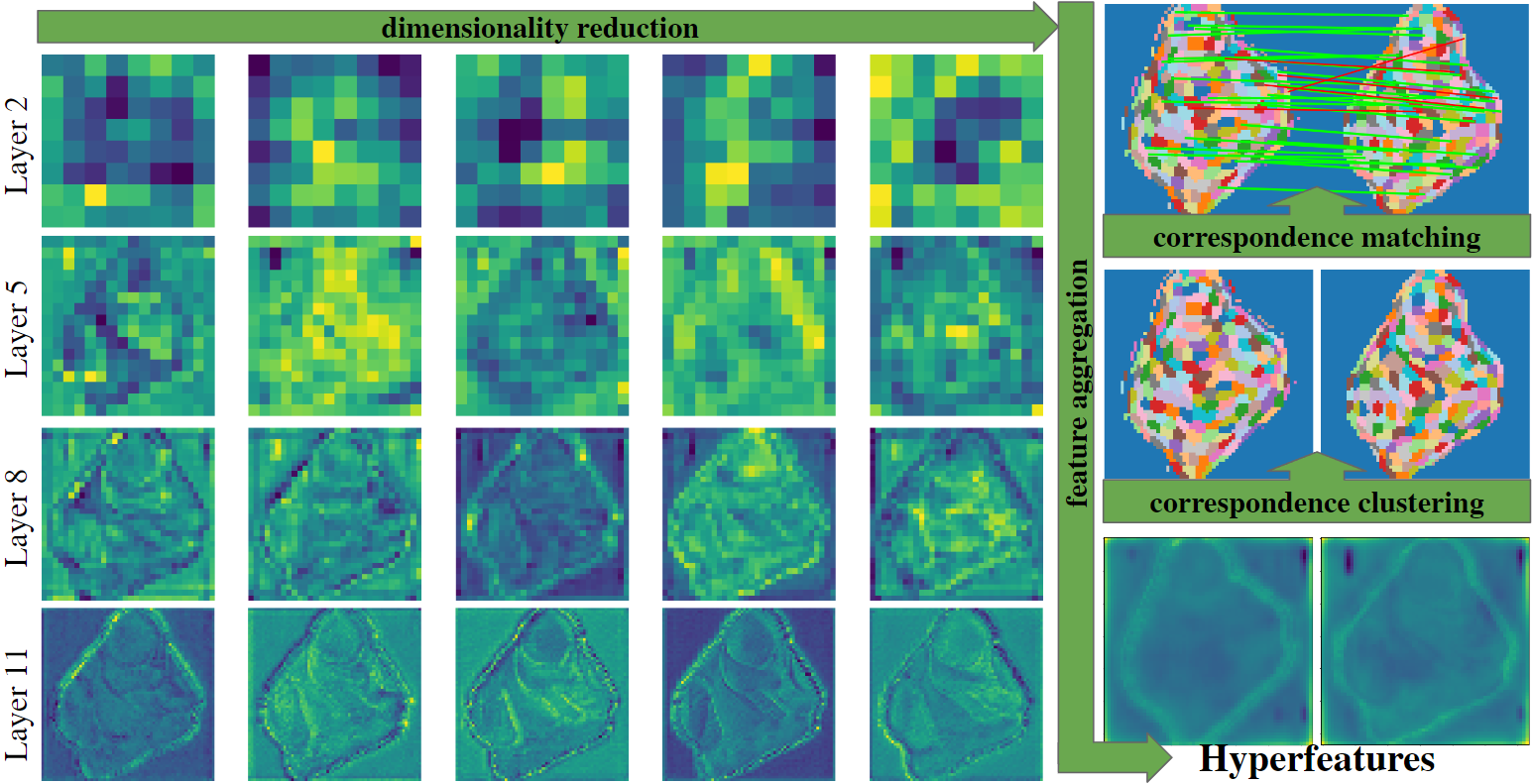}
	\caption{\textbf{Semantic correspondence estimation} First, the query and template features are co-projected to a lower-dimensional space. Subsequently, corresponding clusters are created using cosine similarity. Ultimately, features are matched within corresponding clusters and refined to sub-pixel accuracy.}
	\label{fig:corr}
\end{figure}

In order to retrieve geometric correspondences between the query and the matched template image, mutual semantic correspondences are estimated. 
These correspondences are estimated from hyperfeatures aggregated over multiple feature maps of the U-net's decoder~\cite{zhang2023tale}.
For this, the involved feature maps are projected to a lower dimensional feature space, aggregated, and ultimately clustered to only allow matches between respective clusters. 

The selection of the layer from which to extract features from the backbone is of significant importance, as it influences the semantics that are captured. 
Earlier feature maps of the decoder capture global context and overall structure, whereas later ones focus on details and textures.
The authors of~\cite{zhang2023tale} demonstrate that an aggregation of the feature maps extracted with layer two, five, and eight of SD's backbone enhances the task of correspondence estimation on scene level.
In contrast to their findings, our results indicate that object-level correspondence estimation necessitates a finer texture and geometric detail reasoning than scene-level problems.
Consequently, the incorporation of layer $11$, which is leading to an improved matching accuracy for the task at hand, is intuitive. 

\noindent\textbf{Feature co-projection}
Simple concatenation of feature maps to hyperfeatures results in the formation of high-dimensional features spaces, which in turn results in high computational complexity when these are matched.
Consequently, Principle Component Analysis (PCA) clustering is performed on the concatenated hyperfeatures~\cite{zhang2023tale}:

\begin{equation}
\hat{e}^{l}_{I_q}, \hat{e}^{l}_{I_m} = (e^l(I_q) \cup e^l(I_m)) \cdot V \quad \text{with} \quad U, S, V = PCA(e(I_q) \cup e(I_m)) \quad \text{where} \quad l = 2, 5, 8, 11
\end{equation} \label{gl:co_pca}

\noindent with $\hat{e}^{l}_{I_q}$ and $\hat{e}^{l}_{I_m}$ are obtained with an equal partition along the feature dimension of the dimensionality-reduced union of the feature maps per layer.
Subsequently, feature maps are bilinearly upsampled to the resolution of the finest feature map $L=11$ and concatenation along the feature dimension.

\noindent\textbf{Cluster-wise correspondence estimation}
We observe that a considerable number of mismatches occur when semantic correspondences are estimated in a manner that is independent of their location within the object mask.
Therefore, we estimate correspondences only within corresponding clusters, since this largely alleviates this issue.
K-means clustering is employed to match clusters between the estimated feature embeddings $\hat{e}_{I_q}$ and $\hat{e}_{I_m}$ using the cluster-wise cosine similarity.
Subsequently, RANSAC~\cite{fischler1981random}-based fundamental matrix estimation between the clusters of $\hat{e}_{I_q}$ and $\hat{e}_{I_m}$ is employed to additionally discard mismatched clusters.
The co-projection, as outlined in Equation~\ref{gl:co_pca}, represents a fundamental step for obtaining high cluster similarities between the feature embeddings of $I_q$ and $I_m$, and therefore enables cluster-wise correspondence estimation.

\noindent\textbf{Sub-pixel accurate correspondences}
The objective of correspondence estimation is to derive geometric correspondences between the object's mesh prior and $I_q$.
We observe that low-resolution object location priors, i.e. crops of objects that are small or situated at a distance from the camera, result in aliasing and inaccuracies in correspondence matching. 
This phenomenon is further amplified by the relatively low resolution of the highest resolved feature map, $L11$, which is only one quarter of the resolution of $I_q's$.
Consequently, we interpolate correspondence locations based on their local neighborhood for the purpose of geometric correspondence retrieval.

\begin{equation}
\hat{i}, \hat{j} = round \biggl( \frac{\sum_{0}^{i} cos\_sim (\hat{e}^{i,j}_{I_q}, \hat{e}^{i,j}_{I_m})}{i} \biggr), round \biggl( \frac{\sum_{0}^{j} cos\_sim (\hat{e}^{i,j}_{I_q}, \hat{e}^{i,j}_{I_m})}{j} \biggr) \quad \text{where} \quad cos\_sim = \frac{A \cdot B}{\|A\| \|B\|}
\end{equation} \label{gl:sub_pixel}

\noindent where $i$ and $j$ are the indices within the local neighborhood of matched correspondences. For our experiments we use a kernel size of $3$.

%%% Pose estimation
\subsection{6DoF Pose Retrieval}
The established correspondences between $I_q$ and $I_m$ are employed for the purpose of retrieving the geometric correspondences. 
The image locations of the template's correspondences are converted to three-dimensional geometric correspondences using the strategy of~\cite{park2019pix2pose}.
The geometric correspondences are transferred to the query image using the estimated correspondences,
by projecting them onto the object coordinate frame using the template pose.
The rigid $6DoF$ pose is retrieved using the RANSAC~\cite{fischler1981random} enhanced EP\textit{n}P algorithm~\cite{lepetit2009epnp}.

%%%%%%%%%%%%%%%%%%%%%%%%
%%%%%%%%%%%%%%%%%%%%%
% experimental setup
%%%%%%%%%%%%%%%%%%%%
%%%%%%%%%%%%%%%%%%%%
\section{Experiments}

This section presents a comparison of zero-shot object pose estimation with LDM and ViT. 
The experiments are complemented with results on the accuracy of object pose estimation for different state-of-the-art methods for zero-shot or single-shot inference.
Experiments are performed on three standard datasets, which are also part of the core datasets of the Benchmark for 6D Object Pose Estimation~\cite{hodan2024bop}.
Furthermore, a thorough analysis of the influence of the multiple hyperparameters of SD and the correspondence matching and geometric coordinate retrieval is provided.

%%%%%%%%%%%%%%%%%%%%%
% experimental setup
%%%%%%%%%%%%%%%%%%%%
\subsection{Experimental Setup}

Stable Diffusion\footnote{https://github.com/CompVis/stable-diffusion} is used with U-net as the backbone. 
Pre-training was performed on a subset of LAION-5B~\cite{schuhmann2022laion}.
The SD backbone is used with $50$ timesteps and an embedding dimension of $64$.
DZOP is used with an input image size of $128 \times 128$, $200$ correspondence clusters, and the top $10$ correspondences are estimated between matched clusters.
For geometric correspondence generation and matching, we adopt the approach of~\cite{park2019pix2pose}, as done by ZS6D.

\noindent\textbf{Baseline Method}
The presented approach is based on ZS6D~\cite{ausserlechner2023zs6d}.
For a fair direct comparison between ViT and LDM features we modify the feature extraction stage.
The features are used for template matching and for correspondence estimation. 
The rest of the pipeline, using CNOS to generate object-level priors, geometric correspondence retrieval, and P\textit{n}P for pose retrieval, is used as in ZS6D.
ZS6D uses Dino~\cite{caron2023emerging} as feature extractor.

\noindent\textbf{Datasets}
The datasets used in this work are LMO\cite{brachmann2014learning}, YCBV~\cite{xiang2018posecnn} and TLESS~\cite{hodan2017tless}. 
The LMO dataset is a benchmark for evaluating object pose estimation methods for their robustness to occlusion.  
The particular challenge is dense clutter and the resulting strong occlusion, which increases the difficulty of pose estimation. 
The TLESS dataset contains textureless objects, which are problematic for many recognition algorithms as they often rely on texture features to determine the pose of objects. 
The YCBV dataset contains everyday objects and is a common benchmark for challenging illumination.
Templates are taken from~\cite{nguyen2022templates}, in the case of LMO and TLESS, or generated according to their protocol, in the case of YCBV. 
For comparison, we use exactly the same $300$ templates per object as used by ZS6D.

\noindent\textbf{Metrics}
For comparing $6DoF$ poses we employ the Average Recall (AR)~\cite{hodan2024bop} and for $3D$ template matching the Acc15~\cite{nguyen2022templates}.
The AR is calculated from the Visual Surface Discrepancy (VSD), the Maximum Symmetry-Aware Surface Distance (MSSD), and the Maximum Symmetry-Aware Projection Distance (MSPD):

\begin{multline}\label{eq:VSD}
    e_{VSD} = \underset{p\in \hat{V} \cup V}{avg} \begin{cases} 
    0 &\text{if $p \in \hat{V} \cap V \wedge | \hat{D}(p) - D (p)| < \tau$ }, \\
    1 &\text{otherwise}
    \end{cases} \\
    e_{MSSD} = \underset{s \in S_{s}}{min}\ \underset{m \in M_{s}}{max} ||\hat{P}m - Ps||_{2}, \\
    e_{MSPD} = \underset{s \in S_{s}}{min}\ \underset{m \in M_{s}}{max} ||proj_{3D\to2D}(\hat{P}m) - proj_{3D\to2D}(Psm)||_{2} \\
\end{multline}

\noindent where $\hat{V}$ and $V$ are sets of image pixels; $\hat{D}$ and $D$ are distance maps and $\tau$ is a misalignment tolerance. Distance maps are rendered and compared to the distance map of the test image to derive $\hat{V}$ and $V$.
The indexes $S_{s}$ is a set of symmetry transformations that depend on the visual ambiguities of the object mesh. $M_{s}$ is a subset of the mesh vertices and $proj_{3D\to2D}(.)$ denotes the projection to the image space.
The $AR = (AR^{\theta}_{VSD} + AR_{MSSD} + AR_{MSPD})/3$ is the average recall over the metric.
Where is of the metrics are an average recall over different error thresholds of $\theta$.

The Acc15 quantifies the amount of samples below $15^\circ$ deviation from the ground truth:

\begin{equation}
Acc15 =  \underset{k\in K}{avg} \begin{cases} 
    1 &\text{if} \cos^{-1} \left( \frac{\vec{R^{k}_{q}} \cdot \vec{R^{k}_{m}}}{|\vec{R^{k}_{q}}| |\vec{R^{k}_{m}}|} \right) < 15^\circ, \\
    0 &\text{otherwise}
    \end{cases} \\
\end{equation} \label{gl:acc15}

\noindent where $R_{q}$ and $R_{m}$ are the ground truth rotation of the query and the matched template, respectively.
The variable $K$ denotes the set of individual object instances in the test dataset.

%%%%%%%%%%%%%%%%%%%%%
% Comp sota
%%%%%%%%%%%%%%%%%%%%
\subsection{Comparison with Pose Estimation Methods}

This section provides a comparison with the state of the art for monocular single-shot approaches.
Table~\ref{tab:single-shot} reports AR values for different monocular pose estimation approaches that take single images as input, so-called single-shot methods. 
We report the AR for all methods presented and no pose refinement is applied.
The top block (a) presents instance-level approaches, i.e. those trained on specific object instances and is primarily intended to providing a few examples for reference, while the middle (b) and bottom (c) blocks list zero-shot methods.
%A comparison of DZOP with the baseline method, ZS6D, shows AR improvements on all three datasets. 

\begin{table}[h!]
    \centering
    \begin{tabularx}{15.7cm}{>{\centering\arraybackslash}p{0.5cm}p{3.5cm}>{\centering\arraybackslash}p{4cm}>{\centering\arraybackslash}p{1.7cm}>{\centering\arraybackslash}p{1.7cm}>{\centering\arraybackslash}p{1.7cm}}
    %\begin{tabular}{l|c|c|c|c|c|c}
        & \textbf{Method} & \textbf{Training} & \textbf{LMO} & \textbf{YCBV} & \textbf{TLESS} \\
        \hline
        \textbf{a} & CosyPose \cite{labbe2020cosypose} & instance-level & 0.536 & 0.333 & 0.520 \\
        & GDRnet \cite{wang2021gdr} & instance-level & 0.672 & 0.755 & 0.512 \\ 
        & HiPose \cite{lin2024hipose} & instance-level & 0.799 & 0.907 & 0.833 \\ \hline
        \textbf{b} & MegaPose \cite{labbé2022megapose6dposeestimation} & supervised zero-shot & 0.187 & 0.139 & 0.197 \\
        & OSOP \cite{shugurov2022osopmultistageshotobject} & supervised zero-shot & 0.274 & 0.296 & \textbf{0.403} \\ 
        %OSOP \cite{shugurov2022osopmultistageshotobject} &  $\times$  & 0.312 & \underline{0.332} & - \\ \hdashline
        & GigaPose \cite{nguyen2024gigapose} & supervised zero-shot &  0.299 & 0.290 & 0.273 \\ \hline
        \textbf{c} & FoundPose & unsupervised zero-shot & \textbf{0.396} & \textbf{0.452} & 0.338 \\ \cdashline{2-6}
        & ZS6D \cite{ausserlechner2023zs6d}  & unsupervised zero-shot  & 0.298 & 0.324 & 0.210 \\
        & \textbf{DZOP} (\cite{ausserlechner2023zs6d}+ours) & unsupervised zero-shot  & \underline{0.328} & \underline{0.365} & \underline{0.267} \\
    \end{tabularx}
    \caption{\textbf{Comparison of single-shot monocular object pose estimation comparison} Listed are three different types of single-shot pose estimation approaches: a) instance-level approaches trained offline, b) zero-shot approaches where the feature extractor is fine-tuned to match object observations to rendered views of the respective objects, and c) zero-shot approaches that use VFMs as the feature extractor without fine-tuning.}
    \label{tab:single-shot}
\end{table}

\noindent\textbf{DZOP versus ZS6D}
The underlined values in Figure~\ref{tab:single-shot} indicate the highest AR compared to the ZS6D baseline.
Comparison to Dino~\cite{caron2023emerging} as feature extractor, SD improves the accuracy of zero-shot pose estimation for scenarios with background clutter (LMO), strong illumination changes (YCBV), with symmetric and textureless objects (TLESS), and object occlusion, which is present in all benchmark scenarios.

\noindent\textbf{DZOP versus recent pose estimation approches}
The bold values in Figure~\ref{tab:single-shot} indicate the overall highest zero-shot AR.
Compared to supervised zero-shot, i.e. methods that fine-tune feature extractors for matching images of hand-sized objects on such objects, the proposed method improves on all tested scenarios, except on TLESS. 
OSOP and GigaPose report higher AR on TLESS.
OSOP reports the overall highest AR on TLESS.
The very recently published method FoundPose achieves higher AR than DZOP on all three datasets.
They use a more sophisticated template matching strategy that includes bag-of-words descriptors from Dinov2~\cite{oquab2023dinov2} features, and $800$ templates compared to the $300$ templates used by ZS6D and DZOP.
In addition, poses are estimated for the five best matches, and the best pose is chosen based on the re-projection error of P\textit{n}P, compared to ZS6D's and DZOP's strategy of using only the single matched template to estimate the pose. 
This comparison suggests that Dinov2 features may be superior to Dino and SD features for zero-shot object pose estimation.
However, unlike Dino which uses the classification token, ZS6D uses the positional tokens as used by Dinov2, leading to a significant improvement in the accuracy of correspondence estimation.
Consequently, this observation highlights the need for continuous re-evaluation of recently proposed foundation models on common ground.

%%%%%%%%%%%%%%%
% BIG COMP OF ZS6d and DZOP
%%%%%%%%%%%%%%%
% ZS6D vs DZOP LM
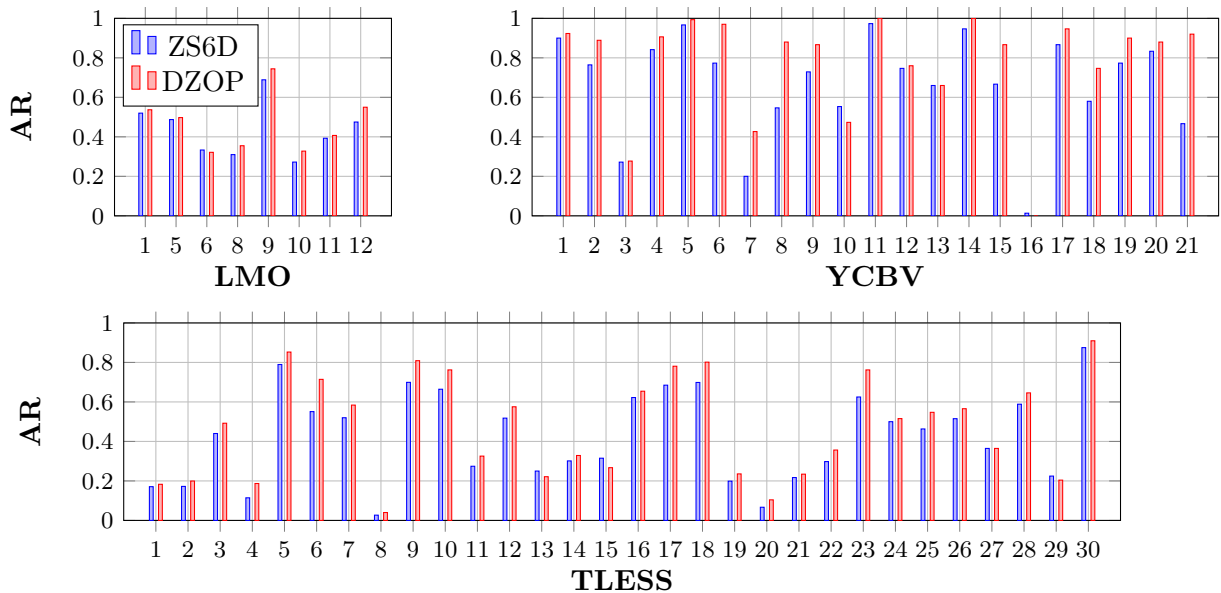
\begin{figure}[ht!]
\centering

\begin{subfigure}[h]{0.32\textwidth}
\centering
\begin{tikzpicture}
\begin{axis}[
ybar,
bar width=0.05cm,
width=0.95\textwidth,
height=4.2cm,
xlabel={\textbf{LMO}},
ylabel={\textbf{AR}},
legend pos=north west,
ytick={0,0.2,0.4,0.6,0.8,1.0},
xtick={1, 2, 3, 4, 5, 6, 7, 8},
xticklabels={1, 5, 6, 8, 9, 10, 11, 12},
xticklabel style={rotate=0},
grid=major,
xmin=0.0,
xmax=9.0,
ymin=0,
ymax=1,
yticklabel style={
/pgf/number format/fixed,
/pgf/number format/precision=2
},
every axis label/.append style={font=\bfseries},
ticklabel style={font=\footnotesize}
]
\addplot coordinates {(1, 0.520) (2, 0.4874) (3, 0.3333) (4, 0.3100) (5, 0.6888) (6, 0.2722) (7, 0.3929) (8, 0.475)};
\addplot coordinates {(1, 0.5371) (2, 0.4975) (3, 0.3216) (4, 0.3550) (5, 0.7444) (6, 0.3277) (7, 0.4071) (8, 0.550)};
\legend{ZS6D, DZOP}
\end{axis}
\end{tikzpicture}
\end{subfigure}
\hspace{0.001cm}
\begin{subfigure}[h]{0.65\textwidth}
\centering
\begin{tikzpicture}
\begin{axis}[
    ybar,
    bar width=0.05cm,
    width=0.95\textwidth,
    height=4.2cm,
    xlabel={\textbf{YCBV}},
    legend pos=north west,
    ytick={0,0.2,0.4,0.6,0.8,1.0},
    xtick={1, 2, 3, 4, 5, 6, 7, 8, 9, 10, 11, 12, 13, 14, 15, 16, 17, 18, 19, 20, 21},
    xticklabel style={rotate=0},
    grid=major,
    xmin=0.0,
    xmax=22.0,
    ymin=0,
    ymax=1.0,
    yticklabel style={
        /pgf/number format/fixed,
        /pgf/number format/precision=2
    },
    every axis label/.append style={font=\bfseries},
    ticklabel style={font=\footnotesize}
]
\addplot coordinates {(1, 0.900) (2, 0.7644) (3, 0.2720) (4, 0.8415) (5, 0.9667) (6, 0.7733) (7, 0.2000) (8, 0.5467) (9, 0.7289) (10, 0.5533) (11, 0.9733) (12, 0.7467) (13, 0.6600) (14, 0.9467) (15, 0.6667) (16, 0.0133) (17, 0.8667) (18, 0.5800) (19, 0.7733) (20, 0.8333) (21, 0.4667)};
\addplot coordinates {(1, 0.9233) (2, 0.8889) (3, 0.2773) (4, 0.9063) (5, 0.9933) (6, 0.9700) (7, 0.4267) (8, 0.8800) (9, 0.8667) (10, 0.4733) (11, 1.0000) (12, 0.7600) (13, 0.6600) (14, 1.0000) (15, 0.8667) (16, 0.0000) (17, 0.9467) (18, 0.7467) (19, 0.9000) (20, 0.8800) (21, 0.9200)};
\end{axis}
\end{tikzpicture}
\end{subfigure}%
\vspace{0.1cm}
\begin{subfigure}[t]{0.95\textwidth}
\flushleft
\begin{tikzpicture}
\begin{axis}[
ybar,
bar width=0.05cm,
width=0.9\textwidth,
height=4.2cm,
xlabel={\textbf{TLESS}},
ylabel={\textbf{AR}},
legend pos=north west,
ytick={0,0.2,0.4,0.6,0.8,1.0},
xtick={1, 2, 3, 4, 5, 6, 7, 8, 9, 10, 11, 12, 13, 14, 15, 16, 17, 18, 19, 20, 21, 22, 23, 24, 25, 26, 27, 28, 29, 30},
xticklabel style={rotate=0},
grid=major,
xmin=0.0,
xmax=31.0,
ymin=0,
ymax=1.0,
yticklabel style={
/pgf/number format/fixed,
/pgf/number format/precision=2
},
every axis label/.append style={font=\bfseries},
ticklabel style={font=\footnotesize}
]
\addplot coordinates {(1, 0.1709) (2, 0.1723) (3, 0.4401) (4, 0.1145) (5, 0.7895) (6, 0.5510) (7, 0.5200) (8, 0.0267) (9, 0.6992) (10, 0.6643) (11, 0.2743) (12, 0.5180) (13, 0.2500) (14, 0.3014) (15, 0.3151) (16, 0.6223) (17, 0.6849) (18, 0.6986) (19, 0.1989) (20, 0.0667) (21, 0.2171) (22, 0.2979) (23, 0.6250) (24, 0.5000) (25, 0.4632) (26, 0.5152) (27, 0.3646) (28, 0.5885) (29, 0.2245) (30, 0.8750)};
\addplot coordinates {(1, 0.1830) (2, 0.1995) (3, 0.4922) (4, 0.1869) (5, 0.8526) (6, 0.7143) (7, 0.5840) (8, 0.0400) (9, 0.8089) (10, 0.7622) (11, 0.3257) (12, 0.5755) (13, 0.2214) (14, 0.3288) (15, 0.2671) (16, 0.6543) (17, 0.7808) (18, 0.8014) (19, 0.2356) (20, 0.1042) (21, 0.2343) (22, 0.3564) (23, 0.7621) (24, 0.5156) (25, 0.5474) (26, 0.5657) (27, 0.3646) (28, 0.6458) (29, 0.2041) (30, 0.9097)};
%\legend{ZS6D, DZOP}
\end{axis}
\end{tikzpicture}
\end{subfigure}

\caption{\textbf{Per-object AR} Reported are the per object improvements of DZOP in comparison to ZS6D.}
\label{fig:zs6d_dzop}
\end{figure}
%%%%%%%%%%%%%%%
% BIG COMP END
%%%%%%%%%%%%%%%

% Platzhalter für die zukünftigen Tabellen und Grafiken zu TLESS und YCBV
\begin{table}[h!]
\centering
%\begin{tabular}{c|c|c|c}
\begin{tabularx}{13.5cm}{p{3cm}>{\centering\arraybackslash}p{3.0cm}>{\centering\arraybackslash}p{3.0cm}>{\centering\arraybackslash}p{3.0cm}}
\textbf{Dataset} & \textbf{Pose Error} & \textbf{Acc15} & \textbf{AR} \\
\hline
LMO & 11.29\% & 8.61\% & 10.08\% \\ %13,76\% \\
YCBV & 4.89\% & 8.75\% & 12.65\% \\
TLESS & -1.14\% & 1.39\% & 27.14\% \\
\end{tabularx}
\caption{\textbf{LDM versus ViT} The relative improvements of DZOP compared to ZS6D are reported for template matching (pose error and Acc15) and $6DoF$ pose estimation (AR).}
\label{tab:mean_comp}
\end{table}

%%%%%%%%%%%%%%% AR thresholds
% ZS6D vs DZOP LM
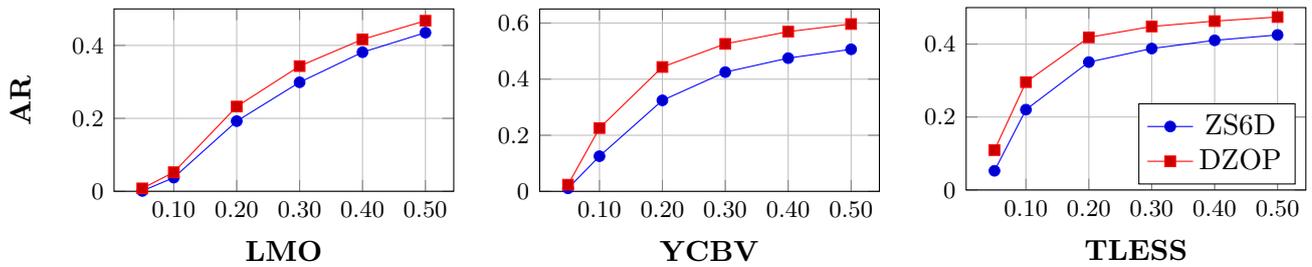
\begin{figure}[h!]
\begin{subfigure}[h]{0.32\textwidth}
\flushleft
\begin{tikzpicture}
\hspace{-0.5cm}
\begin{axis}[
width=1.1\textwidth,
height=4.0cm,
xlabel={\textbf{LMO}},
ylabel={\textbf{AR}},
xtick={0.100, 0.200,0.300,0.400, 0.500},
xticklabels={0.10, 0.20, 0.30, 0.40, 0.50},
ymin=0,
ymax=0.5,
grid=major,
yticklabel style={
/pgf/number format/fixed,
/pgf/number format/precision=3
},
every axis label/.append style={font=\bfseries},
ticklabel style={font=\footnotesize}
]
\addplot coordinates {(0.500, 0.4349) (0.4,0.3813 )(0.300, 0.2989) (0.2,0.19238754325259516 )(0.100, 0.0376)(0.05, 0.001)};
\addplot coordinates {(0.500, 0.4676) (0.4, 0.4166 )(0.300, 0.3432) (0.2, 0.2324)(0.100, 0.0524)(0.05, 0.008)};
\end{axis}
\end{tikzpicture}
\end{subfigure}
\hspace{0.001cm}
\begin{subfigure}[h]{0.32\textwidth}
\centering
\begin{tikzpicture}
\begin{axis}[
width=1.1\textwidth,
height=4.0cm,
xlabel={\textbf{YCBV}},
legend pos=south east,
xtick={0.10, 0.20, 0.30, 0.40, 0.50},
xticklabels={0.10, 0.20, 0.30, 0.40, 0.50},
ymin=0,
ymax=0.65,
grid=major,
yticklabel style={
/pgf/number format/fixed,
/pgf/number format/precision=3
},
every axis label/.append style={font=\bfseries},
ticklabel style={font=\footnotesize}
]
\addplot coordinates {(0.500, 0.5064) (0.400, 0.4749) (0.300, 0.4250) (0.200, 0.3244) (0.100, 0.1254) (0.050, 0.0108)};
\addplot coordinates {(0.500, 0.5965) (0.400, 0.5693) (0.300, 0.5257) (0.200, 0.4433) (0.100, 0.2256) (0.050, 0.0238)};
\end{axis}
\end{tikzpicture}
\end{subfigure}
\hspace{-0.3cm}
\begin{subfigure}[h]{0.32\textwidth}
\flushright
\begin{tikzpicture}
\begin{axis}[
width=1.1\textwidth,
height=4.0cm,
xlabel={\textbf{TLESS}},
legend pos=south east,
xtick={0.10, 0.20,0.30,0.40, 0.50},
xticklabels={0.10, 0.20, 0.30, 0.40, 0.50},
ymin=0,
ymax=0.5,
grid=major,
yticklabel style={
/pgf/number format/fixed,
/pgf/number format/precision=3
},
every axis label/.append style={font=\bfseries},
ticklabel style={font=\footnotesize}
]
\addplot coordinates {(0.500, 0.4250)(0.4, 0.410252)(0.300, 0.3878) (0.2,0.35056 )(0.100, 0.2202)(0.05,0.0526 )};
\addplot coordinates {(0.500, 0.4741)(0.4,0.4630 ) (0.300, 0.4482) (0.2, 0.4184)(0.100, 0.2956)(0.05,0.1093 )};
\legend{ZS6D, DZOP}
\end{axis}
\end{tikzpicture}
\end{subfigure}
\caption{\textbf{AR over error tolerance threshold} Presented are the AR values for different upper bounds of $\theta$.}
\label{fig:AR_thres}
\end{figure}

\subsection{Stable Diffusion versus Dino}
Figure~\ref{fig:zs6d_dzop} shows the AR per object for LMO, YCBV, and TLESS.
On LMO, there is a consistent increase of DZOP compared to ZS6D, resulting in an overall increase of $10.08\%$, see Table~\ref{tab:mean_comp}.
On YCBV, DZOP improves over ZS6D for all objects, except object 10, the \textit{banana}, and 16, the \textit{wood block}.
The pose estimation of object 16 results in an AR close to zero.
This object is inherently difficult due to the multiple discrete symmetries and the difficult natural wood texture.
The average AR increase of DZOP is $12.65\%$.
On TLESS, DZOP improves by an average of $27.14\%$. Improvements are achieved on all objects except from object 29, an object with uniform surface and few geometry cues. 
For object 8 and 20, DZOP and ZS6D report very low AR.
In general, both methods struggle with object symmetries, which is intuitive since they are trained for image flip invariance.
Newer VFMs may alleviate this problem by striving for viewpoint equivariance~\cite{zhang2024telling}.

Table~\ref{tab:mean_comp} shows the relative changes of DZOP compared to ZS6D.  
The pose error indicates the relative cumulative deviation of the matched template's rotation from the ground truth object rotation in degrees. 
DZOP improves significantly on LMO and YCBV. 
Conversely, ZS6D shows more accurate template matching on TLESS, with DZOP resulting in the greatest improvement in AR across all datasets.
Comparing the Acc15, the amount of matched templates below $15^\circ$ rotation deviation, DZOP improves on all three datasets. 
This indicates that DZOP retrieves templates that are closer to the viewpoint of the query image.

Figure~\ref{fig:AR_thres} compares DZOP and ZS6D under different settings for the upper bound of the error tolerance threshold $\theta$.
It can be seen that DZOP also improves even for very low threshold values such as $0.05$. 
In particular, DZOP significantly improves for strong illumination changes for the textured object of YCBV, and the textureless objects of TLESS.
This implies that SD features are much better suited to achieve highly accurate pose estimation in low error tolerance scenarios.
%This means that SD features are much better suited for obtaining highly accurate pose estimates for scenarios where the tolerance for error is low.

\begin{figure}[t!]
	\centering
	\includegraphics[width=.9\columnwidth]{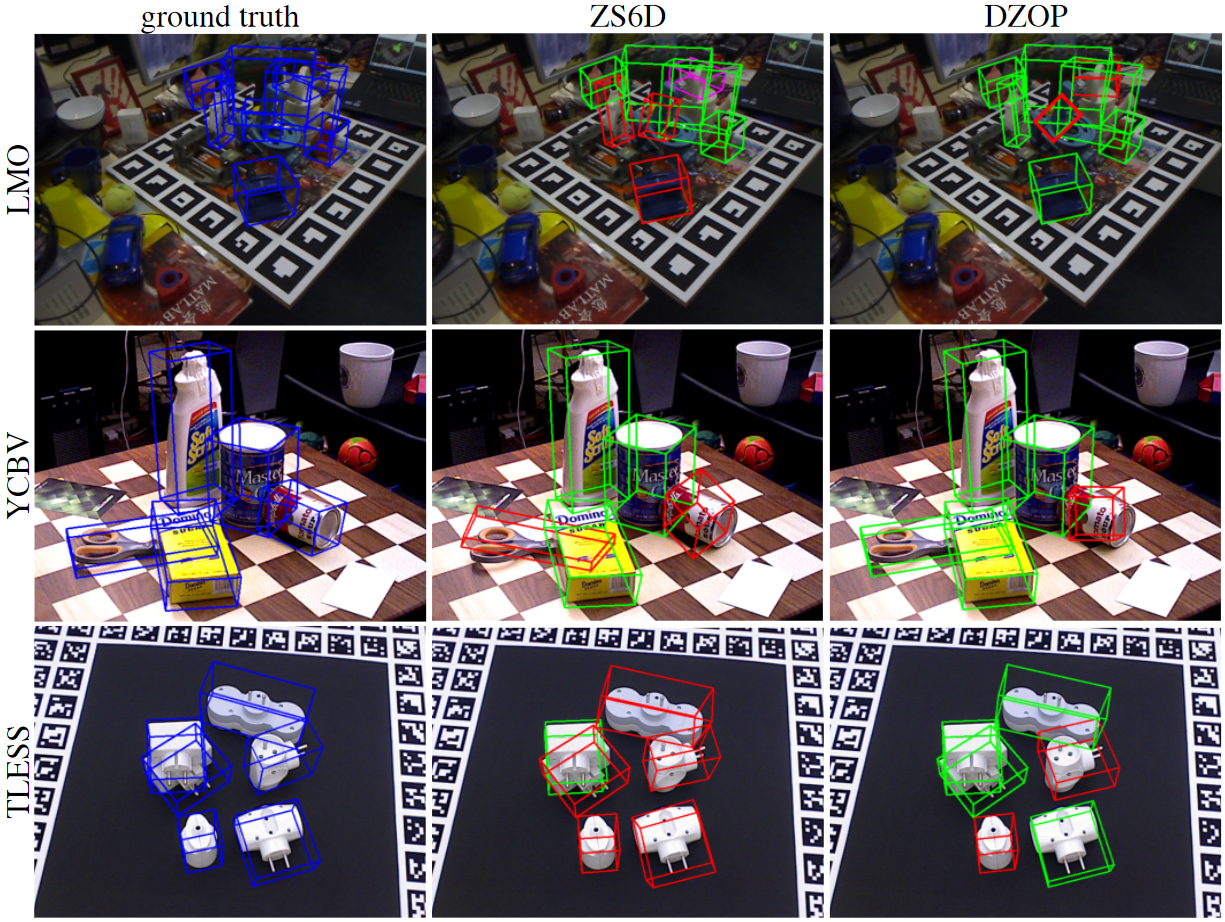}
	\caption{\textbf{Qualitative results} Exemple pose estimates from DZOP and ZS6D. Blue, green, and red boxes indicate ground truth, correct, and incorrect pose estimates respectively.}
	\label{fig:qual_results}
\end{figure}

%deepld until here

Figure~\ref{fig:qual_results} visually compares sample pose estimates from DZOP and ZS6D.
Blue, green, and red boxes indicate ground truth, correct and incorrect pose estimates according to the AR.
On all three datasets DZOP provides more accurate poses.
In particular, the pose estimation of occluded objects improves when using SD as feature extractor, as can be seen on the cat of LMO; in the background of the top row images, and the stacked socket distributors; leftmost object in the bottom row images.
The lower socket distributor is largely occluded, but DZOP's pose estimate is correct according to its AR score. 
On LMO's highly occluded cat ZS6D fails to generate any valid geometric correspondences, while DZOP produces a incorrect, but a still valid pose close to the ground truth.

%%%%%%%%%%%%%%%%%%%%%%%%%
%%%%%%%%%%%%%%%%%%%%%
% Ablations
%%%%%%%%%%%%%%%%%%%%
%%%%%%%%%%%%%%%%%%%%%%%%
\subsection{Ablations}

This section evaluates the method's layout and the hyperparameter choices. 
All ablations are performed using the ground truth bounding boxes as image location prior. 

\noindent\textbf{Correspondence estimation}
Figure~\ref{tab:corr_est} shows the influence of the correspondence matching functions on the AR and runtime, evaluated on LMO.
Runtimes are generated using an NVIDIA GeForce RTX 4070.
Using simple feature concatenation without feature co-projection drastically reduces the AR.
This phenomenon can be explained by the independence of the query and template features, as they are not projected onto the same orthogonal feature space.
Correspondence clustering slightly improves the performance, but again significantly reduces the runtime due to the reduction of the corresponding matching complexity.
Finally, sub-pixel accurate correspondence estimation has little overall impact on the datasets, but improves pose estimation of small objects, such as LMO's \textit{ape}, and objects that are typically far away from the camera, such as LMO's \textit{cat}, shown in the top row of Figure~\ref{fig:qual_results}.

\begin{table}[h!]
    \centering
    \begin{tabularx}{15.7cm}{>{\centering\arraybackslash}p{3.0cm}>{\centering\arraybackslash}p{3.0cm}>{\centering\arraybackslash}p{3.0cm}>{\centering\arraybackslash}p{2.5cm}>{\centering\arraybackslash}p{2.5cm}}
        \textbf{Feature} & \textbf{Correspondence} & \textbf{Sub-pixel} & \textbf{AR} & \textbf{Time [s]}\\ 
        \textbf{co-projection} & \textbf{clustering} & \textbf{accuracy} &  & \\ \hline
        % ZS6D & & & 0.368 \\
        $\times$ & \checkmark & $\times$ & 0.149 & 12.91 \\ 
        \checkmark & $\times$ & $\times$ & 0.427 & 7.18 \\
        \checkmark & \checkmark & $\times$ & 0.432 & 5.29 \\
        \checkmark & \checkmark & \checkmark & 0.433 & 5.29 \\
    \end{tabularx}
    \caption{\textbf{Correspondence estimation} Reported is the impact of each correspondence matching step on the AR and runtime.}
    \label{tab:corr_est}
\end{table}

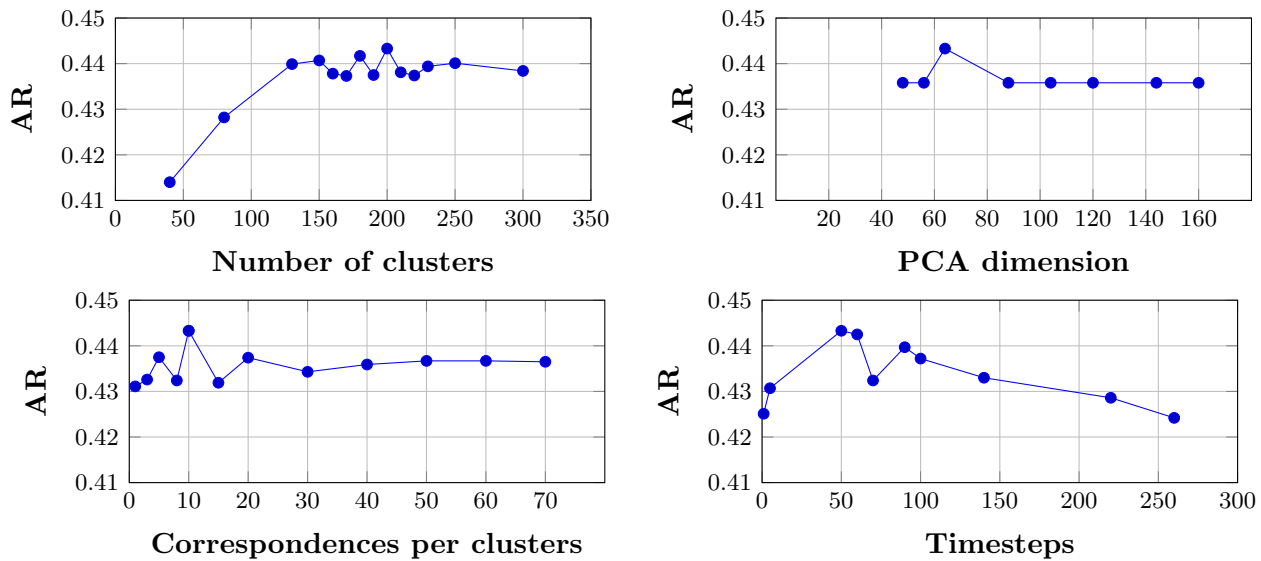
\begin{figure}[t!]
\centering

\begin{subfigure}[h]{0.48\textwidth}
\centering
\begin{tikzpicture}
\begin{axis}[
width=0.95\textwidth,
height=4.0cm,
xlabel={\textbf{Number of clusters}},
ylabel={\textbf{AR}},
ytick={0.410,0.420,0.430,0.440,0.450},
xtick={0,50,100,150,200,250,300,350},
xticklabel style={rotate=0},
grid=major,
xmin=0,
xmax=350.0,
ymin=0.41,
ymax=0.45,
yticklabel style={
/pgf/number format/fixed,
/pgf/number format/precision=2
},
every axis label/.append style={font=\bfseries},
ticklabel style={font=\footnotesize}
]
\addplot coordinates {(40, 0.4140) (80, 0.4282) (130, 0.4399) (150, 0.4407) (160, 0.4378) (170, 0.4373) (180, 0.4417) (190, 0.4375) (200, 0.4433) (210, 0.4381) (220, 0.4374) (230, 0.4394) (250, 0.4401) (300, 0.4384)};
\end{axis}
\end{tikzpicture}
\end{subfigure}
\hspace{0.001cm}
\begin{subfigure}[h]{0.48\textwidth}
\centering
\begin{tikzpicture}
\begin{axis}[
width=0.95\textwidth,
height=4.0cm,
xlabel={\textbf{PCA dimension}},
ylabel={\textbf{AR}},
ytick={0.410,0.420,0.430,0.440,0.450},
xtick={20,40,60,80,100,120,140,160},
xticklabel style={rotate=0},
grid=major,
xmin=0,
xmax=180.0,
ymin=0.41,
ymax=0.45,
yticklabel style={
/pgf/number format/fixed,
/pgf/number format/precision=2
},
every axis label/.append style={font=\bfseries},
ticklabel style={font=\footnotesize}
]
\addplot coordinates {(48, 0.4358) (56, 0.4358) (64, 0.4433) (88, 0.4358) (104, 0.4358) (120, 0.4358) (144, 0.4358) (160, 0.4358)};
\end{axis}
\end{tikzpicture}
\end{subfigure}
\vspace{0.001cm}
\begin{subfigure}[h]{0.48\textwidth}
\centering
\begin{tikzpicture}
\begin{axis}[
width=0.95\textwidth,
height=4.0cm,
xlabel={\textbf{Correspondences per clusters}},
ylabel={\textbf{AR}},
%legend pos=north west,
ytick={0.410,0.420,0.430,0.440,0.450},
xtick={0,10,20,30,40,50,60,70},
xticklabel style={rotate=0},
grid=major,
xmin=0,
xmax=80.0,
ymin=0.41,
ymax=0.45,
yticklabel style={
/pgf/number format/fixed,
/pgf/number format/precision=2
},
every axis label/.append style={font=\bfseries},
ticklabel style={font=\footnotesize}
]
\addplot coordinates {(1, 0.4311) (3, 0.4326) (5, 0.4375) (8, 0.4324) (10, 0.4433) (10, 0.4433) (15, 0.4319) (20, 0.4374) (30, 0.4343) (40, 0.4359) (50, 0.4367) (60, 0.4367) (70, 0.4365)};
\end{axis}
\end{tikzpicture}
\end{subfigure}
\hspace{0.001cm}
\begin{subfigure}[h]{0.48\textwidth}
\centering
\begin{tikzpicture}
\begin{axis}[
width=0.95\textwidth,
height=4.0cm,
xlabel={\textbf{Timesteps}},
ylabel={\textbf{AR}},
ytick={0.410,0.420,0.430,0.440,0.450},
xtick={0,50,100,150,200,250,300,350},
xticklabel style={rotate=0},
grid=major,
xmin=0,
xmax=300.0,
ymin=0.41,
ymax=0.45,
yticklabel style={
/pgf/number format/fixed,
/pgf/number format/precision=2
},
every axis label/.append style={font=\bfseries},
ticklabel style={font=\footnotesize}
]
\addplot coordinates {(1, 0.4251) (5, 0.4307) (50, 0.4433) (60, 0.4425) (70, 0.4324) (90, 0.4397) (100, 0.4372) (140, 0.4330) (220, 0.4286) (260, 0.4242)};
\end{axis}
\end{tikzpicture}
\end{subfigure}
\caption{\textbf{Hyperparameter ablations} Ablated are the main hyperparameters of DZOP: the number of correspondence clusters, the dimensionality of the hyperfeatures, the number of cluster-wise semantic correspondences used for geometric correspondence retrieval, and the timestep of SD used for feature extraction.}
\label{fig:ablations}
\end{figure}

\noindent\textbf{Hyperparameter ablations}
Figure~\ref{fig:ablations} ablates the choice of hyperparameters on LMO's ground truth detections.
The upper left part shows the effect of the number of surface clusters used for correspondence clustering.
While about $125$ to $250$ clusters lead to a comparably high AR a clear maximum can be seen at $200$ clusters.
The top right plot shows that the feature dimension of the co-projection has little influence, but has a distinguishable peak at $64$.
The estimated optimal number of matched correspondences per cluster, bottom left, is $10$.
In general, however, the influence of this parameter is insignificant.
Bottom right shows the number of SD iterations for feature extraction.
A peak is reached at $50$ iterations, with the AR decreasing as the number of timesteps increases.

%%%%%%%%%%%%%%%%
%%%%%%%%%%%%%%%
% 2x2 table for hyperparameters
%%%%%%%%%%%%%%%
%%%%%%%%%%%%%%%

%%%%%%%%%%%%%%%%%%%%
%%%%%%%%%%%%%%%%%%%%%
%% Conclusion
%%%%%%%%%%%%%%%%%%%
%%%%%%%%%%%%%%%%%%%%%%%%
\section{Discussion}

This work investigated whether Latent Diffusion Models improve zero-shot pose estimation compared to Vision Transformers. 
Our results show that Stable Diffusion is a promising alternative to Dino, especially in scenarios characterized by complex object geometries and occluded objects. 
In direct comparison to the baseline the presented method shows significant improvements in Average Recall on three standard datasets.
A comparison of the presented method with a work that uses language-augmented vision embeddings, and a more recent Vision Transformer for matching and correspondence estimation shows improved pose estimation accuracy.
This indicates that further improvements can be obtained using vision and language as two distinct feature embeddings for zero-shot tasks.

Future work will therefore explore different types of mutual feature sets for zero-shot object pose estimation.
Furthermore, the shift from instance-level to zero-shot object pose estimation highlights the need to advance object pose estimation toward computational efficiency. 
Although zero-shot methods reduce the extensive training time required by instance-level approaches, they partially shift the computational burden toward runtime.
This is particularly to be attributed to the number of templates that are required for estimating accurate poses.
Runtime issues can therefore be solved by overcoming the need for object priors.
A promising candidate for solving both of these challenges are multimodal foundation models. 
The aligned feature embedding spaces of vision and language allow access to object priors via semantics, potentially eliminating the need for predefined templates or meshes.
Furthermore, the semantic knowledge of object attributes potentially provides a basis for reasoning about canonical poses across different objects and categories.

%\printbibliography
\bibliographystyle{unsrt}
\bibliography{blibli}
%\bibliographystyle{cas-model1b-num-names}
%\bibliography{cas-refs}

\end{document}